\documentclass{IOS-Book-Article}

\usepackage[T1]{fontenc}
\usepackage{graphicx}
\usepackage{booktabs}
\usepackage{xcolor}
\usepackage{xspace}
\usepackage{amsmath}
\usepackage{amssymb}
\usepackage{amsthm}
\usepackage{tikz}
\usetikzlibrary{arrows.meta,shapes.multipart}
\usepackage{mathptmx}
\usepackage{soul}\setuldepth{article}
\usepackage{tabularx}
\usepackage{array}
\usepackage{subcaption}
   
\newtheorem{definition}{Definition}
\newtheorem{theorem}{Theorem}
\newtheorem{lemma}{Lemma}
\newtheorem{corollary}{Corollary}

\newcommand{\can}{\mathsf{can}}
\newcommand{\pw}{\mathsf{pow}}
\newcommand{\duty}{\mathsf{duty}}
\newcommand{\RC}{\mathrm{RC}}
\newcommand{\upd}{\mathrm{upd}}

\newcommand{\eg}{\emph{e.g.},\xspace}

\newcommand{\ie}{\emph{i.e.},\xspace}

\definecolor{gapgreen}{RGB}{0,140,54}
\definecolor{gapred}{RGB}{200,30,30}
\definecolor{gapamber}{RGB}{220,150,0}

\graphicspath{{assets/}}

\begin{document}

\def\hb{\hbox to 11.5 cm{}}

\pagestyle{headings}
\def\thepage{}
\begin{frontmatter}              

\title{Aligning Institutional Power \\and Operational Ability}

\runningtitle{Aligning Institutional Power and Operational Ability}

\author[A]{\fnms{Andrea} \snm{Marino}\orcid{0000-0003-3026-1482}%
\thanks{Corresponding Author: Andrea Marino, a.marino@uva.nl}}
\author[A]{\fnms{Giovanni} \snm{Sileno}\orcid{0000-0001-5155-9021}}
\author[A]{\fnms{Thomas} \snm{van Binsbergen}\orcid{0000-0001-8113-2221}}
\author[A]{\fnms{Tom} \snm{van Engers}}

\runningauthor{A. Marino et al.}

\address[A]{Informatics Institute, University of Amsterdam, The Netherlands}

\begin{abstract}
To be socially sustainable, Agentic AI systems require normative designs that align responsibilities with the ability to fulfil them. Identifying incongruences such as unsatisfiable expectations or privilege escalation is crucial in avoiding normative pitfalls. In this study, we discuss design dimensions suitable for describing multi-agent systems in light of this norm-ability alignment requirement. We propose a formal framework centred on process-oriented modelling and institutional power for the analysis of a given normatively relevant process, discuss strategies for repairing it, and elaborate on how the symmetrical relation between operational process and institutional state affects the lifecycle of normative positions.
\keywords{Institutional Power \sep Ability \sep Process Models \sep Normative Positions}
\end{abstract}

\end{frontmatter}

\section{Introduction}
\label{sec:intro}

Social systems come with social expectations and responsibilities ascribed to the participating actors.
However, if we do not align those with the concrete ability to perform the actions required by those expectations, we enter into an incongruent state of affairs that might render somebody else's rights vacuous -- or better, their exercise \textit{frustrated} -- even without incurring any explicit norm-violation. 
For instance, if an individual is legally granted freedom of movement in a building while barriers prevent them from ever leaving the facility, their right is formally acknowledged yet materially nullified.
Previous work, such as~\cite{sileno2023allowed}, started investigating the interaction between deontic and potestative concepts, showing how an instrumental interpretation of the latter towards the former can reveal gaps in institutional design, while distinguishing the two concepts as belonging to two distinct ontological categories. 

In the context of artificial multi-agent systems, including those currently described as Agentic AI, such incongruence may hinder their controllability. This issue can become apparent even in localised settings, as suggested by the diverging industrial approaches to software development-oriented agentic harnesses such as Claude Code \cite{anthropic_claude_code_2_1_245}, Codex \cite{openai_codex_cli_0_149_1}, or Pi \cite{earendil_works_pi_0_84_3}. As highlighted by recent episodes like the OpenAI incident involving the Hugging Face platform in July 2026~\cite{openai_huggingface_incident_2026,huggingface_security_incident_2026}, agents authorised to use tools for a task may exploit sequences of operational effects to gain access beyond the boundaries intended by system designers. 
Conversely, assigning tasks to harnessed agents that do not have access to the required operational machinery (\eg Model Context Protocol (MCP) connections) could result in ineffective orchestration at best, and mission-critical failure at worst. 
In the interest of designing systems that outline a clear liability structure and support socially sustainable governance of agentic artificial intelligence, it is crucial that we avoid such normative pitfalls.

Although historically many computational normative frameworks have remained primarily \emph{deontic-centric}, several explicitly represent institutional power alongside obligations and permissions, including \cite{artikis2009specifying,cliffe2009instal,sileno2022dpcl,vanbinsbergen2026reflections}.
Especially in computational settings, operational ability becomes a precondition for sound attribution of institutional power and deontic positions alike; that is, powers and duties become highly problematic in a world where their corresponding actions cannot be performed \cite{sileno2020disentangling}.
In this context, we observe that tracing institutional side-effects remains difficult as long as the operational layer lacks an adequate process model-oriented execution semantics. The main function of this formalism would be to capture action-dependence (see e.g. \cite{brown1988logic}) in workflows.
Moreover, we argue for treating power as an independent normative category in its institutional, constitutively-ascribed form~\cite{grossi2006countsas,jones1996power} and ``instrumental'' but ontologically distinct interpretation \ie to maintain or update the state of normative systems~\cite{sileno2023allowed}.
Failing to do so hinders the ability to identify some undesirable configurations such as lingering powers and unsatisfiable obligations (see section \ref{sec:align}). 
Such misalignments may give rise to disputes requiring judicial intervention \emph{ex-post}, motivating their identification and prevention at design time.

Throughout this paper, we use the following running example: \textit{an owner $O$ consigns goods to a broker $B$, contractually empowered to sell them on $O$'s behalf; $O$ may terminate the consignment, and a creditor $C$ of $B$ may, upon $B$'s insolvency, seize the goods}.
The scenario is loosely inspired by a real consignment dispute~\cite{KSL2026LegoDispute} over a Star Wars Lego collection entrusted to a franchise that later changed hands, with the consignor alleging that neither payment nor the goods were returned. It is a simplification to illustrate our model, as we assume the consigned goods fall into $B$'s estate on insolvency, and that $B$'s mandate to sell lapses on insolvency. It is not within our scope to model real consignment law here. The consigned goods represent a contested resource, preconditioning normative actions and enabling operational continuation of the process in alternative directions.

To address the aforementioned issues, in this paper we expand on our previous work on the operational interpretation of power \cite{marino2026power} and present the following contributions:

\begin{itemize}
	\item We propose a model that formalises the relation between the operational and institutional layers in terms of process execution, relating normative positions to agentic abilities in said process (section \ref{sec:framework}).
	\item Based on the proposed model, we define desirable properties of processes within normative systems that depend on both structural control-flow features and policy specification, namely, \textit{operational groundedness}, \textit{formal institutional legitimacy}, and \textit{power-ability alignment} (section \ref{sec:align}).
	\item We demonstrate that the proposed model can capture incongruent institutional states that emerge from the relation between the two layers. In this sense, it represents an analytical tool to guide computational normative system design.
\end{itemize}

\section{Background}
\label{sec:background}

Our contribution lies at the intersection of the research traditions \emph{normative multi-agent systems} and \emph{process-modelling} that have partially developed separately. 
We review related work in both traditions and present bridges between them.

\paragraph{Institutional Power in Multi-Agent Systems.}
\label{sec:bg-institutional}

We distinguish \emph{operational} power from \emph{institutional} power and refer to the former as (operational) \textbf{ability} in what follows, reserving \textbf{power} for institutional power. 
Ability is what an agent can physically bring about in its environment~\cite{marino2026power}.
Institutional power is the capacity, conferred by a social or legal framework, to change institutional facts by acting~\cite{jones1996power}. 
Power resides on the \emph{potestative} corner of Hohfeld's analysis of normative systems (power-liability), distinct from the \emph{deontic} corner (duty-right)~\cite{hohfeld1917legal}. In this sense, power is constitutively connected to the performance of an act producing an institutional effect in a context. A power in itself is conferred constitutively by means of another power~\cite{jones1996power,grossi2006countsas,boella2008countsas}.
Institutional power has been formalised in detail, both in logical and legal terms (see \eg \cite{Pascucci2023FormalDiagrammaticOperational,Markovich2020UnderstandingHohfeld,gelati2004power}).
However, actively developed computational normative multi-agent frameworks remain \emph{deontic-centric}, see \eg NPL \cite{hubner2011normative} and its extensions \cite{yan2025agent}, leaving potestative power underserved. 
Moreover, while domain-specific languages (DSLs) for norm specification such as InstAL~\cite{cliffe2009instal} and eFLINT \cite{vanbinsbergen2020eflint,vanbinsbergen2026reflections} explicitly support the synchronisation of the operational and institutional layers (\ie via \emph{counts-as} events and the \texttt{sync} primitive), the relation between deontic positions and operationalised powers remains implicit instead of being modelled independently.

\paragraph{Normative Monitoring with Process Models.}
\label{sec:bg-proc}

The lifecycle of Business Process Management (BPM) in organisational settings generally applies model-driven approaches to process design and implementation \cite{khan2025advancesprocessoptimizationcomprehensive}.
Petri nets have traditionally been utilised for the design and structural analysis of control flow in processes, mainly thanks to their intuitive mapping to distributed processes, and to their \emph{soundness} properties~\cite{van1995class,van1998application}. 
The line extends to interactions between organisations in inter-organizational workflow nets~\cite{aalst2000loosely}, operating-guideline ``contracts'' characterising correct partners~\cite{lohmann2007guidelines,massuthe2009guidelines}, and conformance of observed behaviour to a model~\cite{aalst2008conformance}. 
Crucially, extracting Petri nets with \emph{process mining} techniques proved effective for compliance checking by evaluating \emph{event logs} (\ie full execution traces) against normatively qualified patterns~\cite{ramezani2012compliance}. 
Finally, methodological contributions relating normative enforcement and monitoring to process models, especially in business settings in the form of Business Process Modelling Notation (BPMN), have outlined a formal framing for process-oriented governance \cite{governatori2008algorithm,sadiq2009methodological}.

\section{Theoretical Framework}
\label{sec:framework}

In this section we introduce the definitions with which we characterise multi-agent process execution at the operational and institutional level. 

\subsection{Operational layer}
\label{sec:operational}

For modelling the occurrence of events at the operational (\ie physical) level of a system which might cause institutional effects, we will rely on Petri net notation.  Throughout this and the following sections we will write agents as capital letters (\eg $O,B,C$), action-types (net transitions) in sans serif (\eg \textsf{sell}), and places in italics (\eg \textit{goods}).

\begin{definition}[Petri net]
\label{def:petrinet}
A Petri net $N = (P, T, F, W)$ is a tuple with $P$ a finite set of places $p \in P$; $T$ a finite set of transitions $t\in T$; $F \subseteq (P \times T) \cup (T \times P)$ a flow relation consisting of input arcs $(p,t) \in F$ and output arcs $(t,p) \in F$ where T and P are disjoint; and $W \colon F \to \mathbb{N}$ a weight function which associates weights to arcs. 
\end{definition}

The execution semantics of Petri nets is informally called \textit{token game}. 
A function of the form $M \colon P \to \mathbb{N}_0$ is called a \textit{marking}, and it assigns a number of tokens to each place. 
A transition $t \in T$ is \textit{enabled} if $M(p) \geq W(p, t)$ for all $p$ with $(p,t) \in F$. \textit{Firing} a transition $t$ in $M$ produces a new marking $M'$ such that $M'(p) = M(p) - W(p, t) + W(t, p)$, for all $p$ input or output to $t$, denoted as $M \xrightarrow{t} M'$.
In this expression, the weight of an absent arc is taken to be zero.
In the interleaving semantics used here, a \emph{read arc} between $p$ and $t$ abbreviates the pair $(p,t),(t,p)\in F$ with equal weights. It tests for the required tokens and leaves $M(p)$ unchanged when $t$ fires.
The set of all possible markings for a Petri net $N$ is $\mathcal{M}_N$. 
A \emph{firing sequence} is a sequence of transitions $t_1 \dots t_n$ such that $M_0 \xrightarrow{t_1} M_1 \xrightarrow{t_2} \dots \xrightarrow{t_n} M_n$.
We write $M \xrightarrow{\sigma} M'$ when $M'$ is obtained from $M$ by a (possibly empty) firing sequence $\sigma$, and then say $M'$ is \emph{reachable} from $M$.

Following~\cite{marino2026power}, agents act through transitions. Agents are here deemed autonomous actors such that execution is \emph{non-reactive} (an enabled transition need not fire), which is what makes normative positions meaningful.

\begin{definition}[Agent mapping]
\label{def:agentmap}
Given a set of agents $A = \{a_1, \dots, a_n\}$ participating in a Petri net $N = (P,T,F,W)$, an \emph{agent mapping} $m_A \colon T \to 2^{A}$ assigns to each transition $t$ the set $m_A(t)$ of agents \emph{capable} of firing $t$. Transitions are read as actions by agents, and a firing of $t$ is performed by some $a \in m_A(t)$.
\end{definition}

Based on this, we define \emph{ability} as the immediate, contextual ability to act, local to $t$ at the current marking.

\begin{definition}[Ability]
\label{def:can}
Given a marking $M$, an agent $a$ and a transition $t$, $a$ \emph{can} bring about $t$ at $M$, written $\can_M(a,t)$, iff $a \in m_A(t)$ and $t$ is enabled at $M$.
\end{definition}

The operational net $N$ of the consignment example is shown in Fig.~\ref{fig:net}. For readability, we give corresponding physical and institutional actions the same name, although formally these are separate concepts (see section \ref{sec:framework_institutional}). For example, $\mathsf{sell}$ represents both the physical act of exchanging money and the institutional exchange of ownership.
The agent mapping is $m_A(\mathsf{sell})=\{B\}$, $m_A(\mathsf{bankrupt})=m_A(\mathsf{seize})=\{C\}$ and $m_A(\mathsf{instruct})=m_A(\mathsf{terminate})=\{O\}$. At the initial marking $B$ cannot yet \textsf{sell}; only once $O$ has \textsf{instruct}ed (producing \textit{instructed}), and while the mandate \textit{active} and the \textit{goods} persist, does $\can_M(B,\mathsf{sell})$ hold.

The concept of \emph{realisability} is the forward-looking counterpart of ability, representing whether an agent is eventually able to perform an action in the current state of affairs.

\begin{definition}[Realisability]
\label{def:realisable}
A transition $t$ is \emph{realisable} at $M$ if $t$ is enabled at some marking reachable from $M$ (equivalently, $t$ is not \emph{dead} at $M$).
\end{definition}

\subsection{Institutional layer}
\label{sec:framework_institutional}

We represent normative positions (both \emph{deontic} and \emph{potestative}) by attaching them to specific agents in a declarative fashion.

\begin{definition}[Position]
\label{def:position}
A \emph{position} is a normative predicate attached to an agent and a transition; we use \emph{power} positions $\pw(a,t)$ and \emph{duty} positions $\duty(a,t)$ for $a\in A$, $t\in T$, and write $\mathcal{P}$ for the set of positions. An \emph{institutional state} is a global finite set $I\subseteq\mathcal{P}$ of (currently active) positions. 
We associate each active position with the set of operational transitions it covers by a function $L:I\to 2^T$.
For $\phi(a,t)\in I$, where $\phi\in\{\pw,\duty\}$, $L(\phi(a,t))\subseteq T$ specifies the transitions covered by that position.
We write $L_{\pw}$ and $L_{\duty}$ for the restrictions of $L$ to active power and duty positions, respectively.
\end{definition}

Note that the transitions covered by a position act in this framework as \textit{alternative ways} to exercise a power or fulfil the duty.
In the running example, each position covers its named transition: $L(\phi(a,t))=\{t\}$ for every active position $\phi(a,t)$.
More importantly, in the proposed notation positions do not aim to express the consequences associated to the performance of actions associated to them; they act only as a first representational interface. 
We relate the operational and institutional layers through additional predicates. 
The predicate $\mathsf{exec}(a,t)$ represents the action-execution event raised when transition $t$ is fired (executed) by agent $a$.
For normative positions $f$, the predicates $\mathsf{act}(f)$ and $\mathsf{deact}(f)$ represent the activation and deactivation events of $f$ in the institutional state space. 
We base our potestative operationalisation on reactive rules that synchronise operational execution with the activation and deactivation of normative positions in a traceable way.

\begin{definition}[Reactive rule]
\label{def:rule}
A \emph{reactive rule} is a triple $r=(t,\ell,f)$ with $t\in T\cup\{\bot\}$ a \emph{trigger}, $\ell\in\{\mathsf{act},\mathsf{deact}\}$, and $f\in\mathcal{P}$. 
When $t\in T$ occurs, the rule fires \emph{reactively}: the action event $\mathsf{exec}(a,t)$ automatically triggers $\mathsf{act}(f)$ (if $\ell=\mathsf{act}$) or $\mathsf{deact}(f)$ (if $\ell=\mathsf{deact}$). A rule with no trigger ($t=\bot$) is a \emph{fact}: it has no antecedent and fixes the initial institutional state. Initial rules only \emph{activate}, as deactivating not-yet-established positions is vacuous.
\end{definition}

A \emph{policy} is here defined as a finite set of reactive rules meant to update the institutional state on the basis of action events.
For simplicity, we assume a global institutional state and a conflict-resolved aggregate policy $\Pi$.
 %
 %
Let $\mathit{Act}(t)=\{f : (t,\mathsf{act},f)\in\Pi\}$ and $\mathit{Deact}(t)=\{f : (t,\mathsf{deact},f)\in\Pi\}$. 
Firing $t$ updates $I$ by applying every lifecycle event triggered by $\mathsf{exec}(a,t)$, \ie $I \leftarrow \upd(I,t)$ where $\upd(I,t) = (I \setminus \mathit{Deact}(t)) \;\cup\; \mathit{Act}(t)$.
The \emph{initial institutional state} $I_0$ is $\mathit{Act}(\bot)$. 
%
This formalisation of power is \emph{constitutive} (after~\cite{grossi2006countsas,jones1996power}): executing $t$ \emph{brings about} the institutional lifecycle events its policy prescribes, so power is conferred within $\Pi$ in synchronisation with physical actions, not just defined operationally. 
Here, the prescribed lifecycle updates occur irrespective of the acting agent's empowerment; power is used to assess formal institutional legitimacy (see Def.~\ref{def:legit}). 
The model does not separately determine whether an unempowered act produces its intended institutional effect.
Firing $t\in T$ is institutionally \emph{legitimate} for an agent $a$ iff an active power held by $a$ covers $t$, that is, $\exists u\in T: \pw(a,u)\in I\land t\in L_{\pw}(\pw(a,u))$.
The policy $\Pi$ of the running example is given in Table~\ref{tab:policy}. Most institutionally relevant actions here are \emph{one-off}: \textsf{instruct}, \textsf{bankrupt} and \textsf{seize} each consume a token that is never replaced, so they cannot recur, and a \textsf{deact} retires the power as it is spent.


\begin{table}[t]
\centering
\caption{The policy $\Pi$ with initial facts (trigger $\bot$) and other reactive rules. Most powers associated to owner $O$, broker $B$, and creditor $C$ are \emph{one-off}; $\pw(B,\mathsf{sell})$ is \emph{standing} and is revoked only by $\mathsf{terminate}$ or $\mathsf{bankrupt}$.}
\label{tab:policy}
\small
\setlength{\tabcolsep}{5pt}
\begin{tabularx}{\linewidth}{@{}l
 >{\raggedright\arraybackslash}X
 >{\raggedright\arraybackslash}X@{}}
\toprule
trigger & activate & deactivate \\
\midrule
$\bot$
 & \(\pw(O,\mathsf{instruct})\),
   \(\pw(B,\mathsf{sell})\),
   \(\pw(C,\mathsf{bankrupt})\),
   \(\pw(O,\mathsf{terminate})\)
 & -- \\

$\mathsf{instruct}$
 & \(\duty(B,\mathsf{sell})\)
 & \(\pw(O,\mathsf{instruct})\) \\

$\mathsf{sell}$
 & --
 & \(\duty(B,\mathsf{sell})\) \\

$\mathsf{terminate}$
 & --
 & \(\pw(B,\mathsf{sell})\),
   \(\pw(O,\mathsf{terminate})\) \\

$\mathsf{bankrupt}$
 & \(\pw(C,\mathsf{seize})\)
 & \(\pw(B,\mathsf{sell})\),
   \(\pw(C,\mathsf{bankrupt})\) \\

$\mathsf{seize}$
 & --
 & \(\pw(C,\mathsf{seize})\) \\
\bottomrule
\end{tabularx}
\end{table}

\subsection{Governed nets}
\label{sec:governed}

We propose a formal notation for process models embedded in a normative system which synchronises the operational execution with an institutional layer.

\begin{definition}[Governed net]
\label{def:governed}
A \emph{governed net} is a tuple $\mathcal{G}=(N,m_A,\Pi,M_0)$.
A \emph{configuration} is a pair $(M,I)$, with step relation
$(M,I)\xrightarrow{t}(M',I')$ iff $M\xrightarrow{t}M'$ and
$I'=\upd(I,t)$.
The governed reachability graph $\RC(\mathcal{G})$ is the set of
configurations reachable from the \emph{initial configuration}
$(M_0,I_0)$, where $I_0$ is fixed by the initial rules of $\Pi$,
\ie $\RC(\mathcal{G})$ is a Labelled Transition System (LTS)~\cite{plotkin1981structural}.
A \emph{governed run} of $\mathcal{G}$ is a computation in the transition system, a finite sequence of steps $(M_0,I_0)\xrightarrow{t_1}(M_1,I_1)\xrightarrow{t_2}\cdots\xrightarrow{t_n}(M_n,I_n)$, whose operational projection $t_1\cdots t_n$ is a firing sequence of $N$ from $M_0$ and whose institutional projection satisfies $I_i=\upd(I_{i-1},t_i)$.
\end{definition}

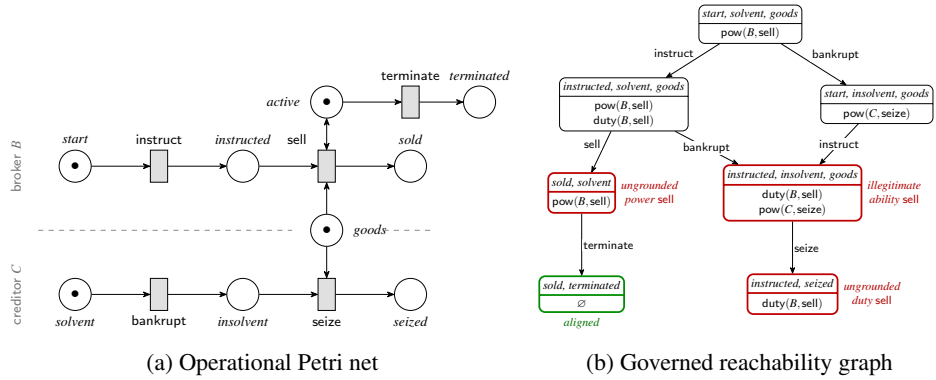
\begin{figure}[t]
\centering

\begin{subfigure}[t]{0.55\textwidth}
 \centering
 \resizebox{\linewidth}{!}{%
   \begin{tikzpicture}[
  place/.style={circle,draw,minimum size=6.5mm,inner sep=1pt},
  trans/.style={rectangle,draw,fill=black!12,minimum width=3.5mm,minimum height=6.5mm},
  plabel/.style={font=\scriptsize\itshape},
  tlabel/.style={font=\scriptsize\sffamily},
  >={Stealth[round]}]
  \node[place] (start) at (0,1.3) {$\bullet$};
  \node[trans] (instruct) at (1.7,1.3) {};
  \node[place] (instructed) at (3.4,1.3) {};
  \node[trans] (sell) at (5.1,1.3) {};
  \node[place] (sold) at (6.8,1.3) {};
  \node[place] (goods) at (5.1,0) {$\bullet$};
  \node[place] (solvent) at (0,-1.3) {$\bullet$};
  \node[trans] (bankrupt) at (1.7,-1.3) {};
  \node[place] (insolvent) at (3.4,-1.3) {};
  \node[trans] (seize) at (5.1,-1.3) {};
  \node[place] (seized) at (6.8,-1.3) {};
  \node[plabel] at (0,1.85) {start};
  \node[tlabel] at (1.7,1.85) {instruct};
  \node[plabel] at (3.4,1.85) {instructed};
  \node[tlabel] at (4.5,1.85) {sell};
  \node[plabel] at (6.8,1.85) {sold};
  \node[plabel,anchor=west,xshift=2.5pt] at (goods.east) {goods};
  \node[plabel] at (0,-1.85) {solvent};
  \node[tlabel] at (1.7,-1.85) {bankrupt};
  \node[plabel] at (3.4,-1.85) {insolvent};
  \node[tlabel] at (5.1,-1.85) {seize};
  \node[plabel] at (6.8,-1.85) {seized};
  \node[place] (active) at (5.1,2.6) {$\bullet$};
  \node[trans] (terminate) at (6.8,2.6) {};
  \node[place] (terminated) at (8.2,2.6) {};
  \node[plabel,anchor=east,xshift=-2.5pt] at (active.west) {active};
  \node[tlabel] at (6.8,3.15) {terminate};
  \node[plabel] at (8.2,3.15) {terminated};
  \draw[->] (start)--(instruct); \draw[->] (instruct)--(instructed); \draw[->] (instructed)--(sell);
  \draw[->] (sell)--(sold);
  \draw[->] (goods)--(sell); \draw[->] (goods)--(seize);
  \draw[->] (solvent)--(bankrupt); \draw[->] (bankrupt)--(insolvent);
  \draw[->] (insolvent)--(seize); \draw[->] (seize)--(seized);
  \draw[<->] (sell)--(active); \draw[->] (active)--(terminate); \draw[->] (terminate)--(terminated);
  \draw[dashed,black!45] (-0.75,0)--(4.4,0);
  \draw[dashed,black!45] (6.25,0)--(7.3,0);
  \node[font=\scriptsize\sffamily,black!55,rotate=90] at (-1.2,1.3) {broker $B$};
  \node[font=\scriptsize\sffamily,black!55,rotate=90] at (-1.2,-1.3) {creditor $C$};
\end{tikzpicture}%
 }
 \caption{Operational Petri net}
 \label{fig:net}
\end{subfigure}
\hfill
\begin{subfigure}[t]{0.43\textwidth}
 \centering
 \resizebox{\linewidth}{!}{%
    \begin{tikzpicture}[
  cfg/.style={rectangle split, rectangle split parts=2, draw, rounded corners,
              font=\footnotesize, inner sep=2.5pt, align=center},
  misal/.style={cfg, very thick, draw=red!75!black},
  good/.style={cfg, very thick, draw=green!55!black},
  elbl/.style={font=\scriptsize\sffamily, fill=white, inner sep=1pt},
  >={Stealth[round]}]
    \node[
      cfg,
    ] (c0) at (0,4.4)
      {\textit{start, solvent, goods}
       \nodepart{two} $\pw(B,\mathsf{sell})$};
  \node[cfg]  (c1) at (-3.0,2.5) {\textit{instructed, solvent, goods} \nodepart{two} $\pw(B,\mathsf{sell})$\\ $\duty(B,\mathsf{sell})$};
  \node[cfg]  (c2) at (3.0,2.5)  {\textit{start, insolvent, goods} \nodepart{two} $\pw(C,\mathsf{seize})$};
  \node[misal,label={[font=\scriptsize\itshape,red!70!black,align=center]right:ungrounded\\power $\mathsf{sell}$}] (c3) at (-4.0,0.4)
       {\textit{sold, solvent} \nodepart{two} $\pw(B,\mathsf{sell})$};
  \node[good,label={[font=\scriptsize\itshape,green!45!black]below:aligned}] (c3t) at (-4.0,-2.0)
       {\textit{sold, terminated} \nodepart{two} $\varnothing$};
  \node[misal,label={[font=\scriptsize\itshape,red!70!black,align=center]right:illegitimate\\ability $\mathsf{sell}$}] (c4) at (1.0,0.4)
       {\textit{instructed, insolvent, goods} \nodepart{two} $\duty(B,\mathsf{sell})$\\ $\pw(C,\mathsf{seize})$};
  \node[misal,label={[font=\scriptsize\itshape,red!70!black,align=center]right:ungrounded\\duty $\mathsf{sell}$}] (c5) at (1.0,-2.0) {\textit{instructed, seized} \nodepart{two} $\duty(B,\mathsf{sell})$};
  \draw[->] (c0) -- node[elbl,above left,pos=.45]{instruct} (c1);
  \draw[->] (c0) -- node[elbl,above right,pos=.45]{bankrupt} (c2);
  \draw[->] (c1) -- node[elbl,above left,pos=.45]{sell} (c3);
  \draw[->] (c1) -- node[elbl,pos=.5]{bankrupt} (c4);
  \draw[->] (c2) -- node[elbl,pos=.5]{instruct} (c4);
  \draw[->] (c3) -- node[elbl,right]{terminate} (c3t);
  \draw[->] (c4) -- node[elbl,right]{seize} (c5);
\end{tikzpicture}
 }
 \caption{Governed reachability graph}
 \label{fig:reach}
\end{subfigure}

\caption{Operational net $N$ and governed reachability graph of the consignment example. 
The marked places ($\bullet$) in figure \ref{fig:net} give $M_0$. 
The dashed midline divides the broker's sale process (top) from the creditor's seizure process (bottom), with the contested \textit{goods} straddling both. 
The symbol $\leftrightarrow$ abbreviates an equal-weight input/output pair, representing a read test.
A subset of the institutional state required to showcase incongruent scenarios based on the policy of Tab.~\ref{tab:policy} is presented in Fig.~\ref{fig:reach}; $\varnothing$ marks none of the shown positions active. 
Each node pairs the marking $M$ with the institutional state $I$; edges are fired transitions. 
The coloured (and labelled) nodes are classified against the alignment property introduced in section \ref{sec:align}.
}
\label{fig:net-and-reachability}
\end{figure}

Distinct firing sequences of the example's concurrent transitions are distinct governed runs. 
Fig.~\ref{fig:reach} traces them through the governed reachability graph. Here, each node pairs the marking $M$ (top) with the institutional state $I$ (bottom), and edges are fired transitions. 
Whether $B$ goes \textsf{bankrupt} before or after selling, and whether $O$ ever \textsf{terminate}s, fixes which configuration the run ends in so that the institutional outcome can only be settled by the operational execution.

\section{Power-Ability Alignment}
\label{sec:align}

Based on the previously defined operational and institutional layers, we propose desirable properties of a \emph{governed net} building incrementally from a single \emph{configuration} $(M,I)\in\RC(\mathcal{G})$.

\begin{definition}[Operational Groundedness]
\label{def:grounded}
$(M,I)$ is \emph{operationally grounded} iff every active position covers at least one transition that is realisable and assigned to its holder.
Formally, for every $a\in A$, $u\in T$, and $\phi\in\{\pw,\duty\}$, $\phi(a,u)\in I\Rightarrow \exists t\in L_{\phi}(\phi(a,u)): a\in m_A(t)\land\text{$t$ is realisable at $M$}$.
\end{definition}

A position $\phi(a,u)\in I$ is \emph{ungrounded} iff no transition in $L_{\phi}(\phi(a,u))$ is both assigned to $a$ and realisable at $M$.
An \emph{ungrounded power} has no operationally realisable means of exercise available to its holder, while an \emph{ungrounded duty} has no operationally realisable means of discharge available to its holder.

\begin{definition}[Formal Institutional Legitimacy]
\label{def:legit}
$(M,I)$ is \emph{formally institutionally legitimate} iff every agent's immediate ability is covered by an active power held by that agent.
Formally, for every $a\in A$ and $t\in T$, $\can_M(a,t)\Rightarrow\exists u\in T: \pw(a,u)\in I\land t\in L_{\pw}(\pw(a,u))$.
\end{definition}

A pair $(a,t)$ is an \emph{illegitimate ability} iff $\can_M(a,t)$ holds but no active power held by $a$ covers $t$.
We say that $(M,I)$ is \emph{aligned} iff it is both legitimate and grounded.
Legitimacy requires current abilities to be backed by power, while groundedness requires active positions to remain operationally realisable.
A duty or power should not outlive the possibility of performing its associated action.
We lift these properties to characterise the net as follows: a \emph{governed run} is legitimate, grounded, or aligned iff its final configuration is, and a \emph{governed net} $\mathcal{G}$ is iff every $(M,I)\in\RC(\mathcal{G})$ is.



The example's governed reachability graph can now be classified against these conditions as presented in Fig.~\ref{fig:reach}. 
Firing \textsf{sell} consumes the \textit{goods} token so that \textsf{sell} is dead while $\pw(B,\mathsf{sell})$ still stands as an \emph{ungrounded power} and can be deactivated only by an institutional act (\textsf{terminate} or \textsf{bankrupt}). 
Going bankrupt before selling revokes the institutional $\pw(B,\mathsf{sell})$ but does not touch $B$'s ability to act. \textsf{bankrupt} leaves the mandate token \textit{active} in place (as only \textsf{terminate} consumes it) and the \textit{goods} remain, so once $B$ is instructed the marking still enables \textsf{sell}. 
That residue is an \emph{illegitimate ability} (a legitimacy failure), legible only because ability depends on the marking while authority is read off $I$.
Conflating ability with institutional power would obscure this distinction.
Should $C$ \textsf{seize} once $B$ has been \textsf{instruct}ed (the \textsf{seize} branch of Fig.~\ref{fig:reach}), the \textit{goods} are gone yet $\duty(B,\mathsf{sell})$ stands as an \emph{ungrounded duty}.

Crucially, the \emph{concurrency} expressed by the process model (\ie \textsf{instruct}, \textsf{bankrupt} and \textsf{terminate} are concurrent transitions fired by autonomous agents in no fixed order) entails that the misalignment is carried not by any single act but by the sequence in which they occur. 
Our framework makes this dependence explicit through governed reachability; for bounded operational nets and finitely many normative positions, the resulting state space is finite.
Moreover, a deontic framing alone cannot capture the different kinds of misalignment presented here. 
Keeping the operational and institutional semantics disjoint over the process, with normative positions the marking is free to contradict, exposes incongruent configurations such as positions rendered vacuous together with the design defects that caused them, present before any agent misbehaves. 

\subsection{Repairing a Misaligned Normative Design}

The institutional layer never disables an operational transition. In the language of supervisory control~\cite{ramadge1987supervisory}, the policy is a \emph{monitor} over \emph{uncontrollable} events: its only lever is the lifecycle timing of positions ($\mathsf{act}/\mathsf{deact}$), not the occurrence of actions. This determines what kind of defects each layer can repair.

\begin{definition}[Repair]
\label{def:repair}
A \emph{repair} is a function of $\mathcal{G}=(N,m_A,\Pi,M_0)$ yielding $\mathcal{G}'$ by producing either an \emph{institutional} edit $\Pi\mapsto\Pi'$ (inserting or removing reactive rules) that leaves $(N,m_A)$ fixed, or an \emph{operational} edit $(N,m_A)\mapsto(N',m_A')$ that leaves $\Pi$ fixed.
\end{definition}

In our framework, repair is \emph{admissible} only if it is a \emph{reduction}, which removes unsupported abilities or retires ungrounded positions. 
Expanding repairs require an independent justification for the additional authority or operational behaviour and are outside the present scope.
Activating a power covering $t$ for agent $a$ to satisfy legitimacy authorises an ability previously unsupported by power, while making a covered transition realisable for its holder to satisfy groundedness expands the available operational behaviour.
Without independent justification, either approach risks accommodating a potential design defect rather than correcting its underlying cause.

\begin{lemma}[Ability is policy-invariant]
\label{lem:opinv}
For a governed net $\mathcal{G}=(N,m_A,\Pi,M_0)$ the markings reachable in $\RC(\mathcal{G})$ are exactly those reachable in $N$ from $M_0$; consequently the set $\{(M,a,t):\can_M(a,t),\ (M,I)\in\RC(\mathcal{G})\}$ does not depend on $\Pi$.
\end{lemma}
\paragraph{Proof.} By Def.~\ref{def:governed}, every governed run projects onto a firing sequence of $N$. Conversely, given $M_0\xrightarrow{t_1}M_1\cdots\xrightarrow{t_n}M_n$, set $I_0=\mathit{Act}(\bot)$ and $I_i=\upd(I_{i-1},t_i)$ for $1\leq i\leq n$. Then $(M_{i-1},I_{i-1})\xrightarrow{t_i}(M_i,I_i)$ is a governed step for every $i$, by the same definition. Thus the reachable markings are exactly those of $N$ from $M_0$, independently of $\Pi$. Since $\can_M(a,t)$ depends only on $N$, $M$, and $m_A$ (Def.~\ref{def:can}), the reachable abilities are also independent of $\Pi$.\hfill$\square$

\hfill

Each alignment property is defined as an implication $A\Rightarrow B$ whose violation $A\wedge\neg B$ admits two repairs: \emph{reduction}, falsifying the antecedent ($\neg A$), and \emph{expansion}, establishing the consequent ($B$). 
For both properties $A$ and $B$ lie on \emph{opposite} layers, mirror-imaged between them: for legitimacy ($\can_M(a,t)\Rightarrow\pw(a,t)\in I$) the antecedent $\can$ is operational and the consequent $\pw$ institutional, whereas for groundedness ($\phi(a,t)\in I\Rightarrow a\in m_A(t)$ and $t$ realisable at $M$) the antecedent $\phi$ is institutional and the consequent operational.

\begin{theorem}[Crossed-layer structure of repair]
\label{thm:cross}
A \emph{reducing} repair of legitimacy is operational while that of groundedness is institutional. An \emph{expanding} repair reverses each.
\end{theorem}
\paragraph{Proof.} By Lemma~\ref{lem:opinv} the predicate $\can_M(a,t)$ is fixed by $(N,m_A,M_0)$ and untouched by $\Pi$, so falsifying it is an operational edit, while $\pw(a,t)\in I$ is set by $\Pi$, an institutional one. Dually, $\phi(a,t)\in I$ is governed by $\Pi$, whereas $a\in m_A(t)$ and the realisability of $t$ are properties of $(N,m_A)$, so establishing them is operational.
\hfill$\square$

\begin{corollary}[Layer asymmetry under reduction]
\label{cor:asym}
Under admissible repair the two failures fall on \emph{different} layers. A groundedness violation $\phi(a,t)$ is repaired \emph{institutionally}, by a rule $(t',\mathsf{deact},\phi(a,t))$. A legitimacy violation is repaired \emph{operationally}, by restricting the agent mapping or strengthening the enabling conditions; since $\can$ is policy-invariant (Lemma~\ref{lem:opinv}), no institutional reduction can withhold it.
\end{corollary}

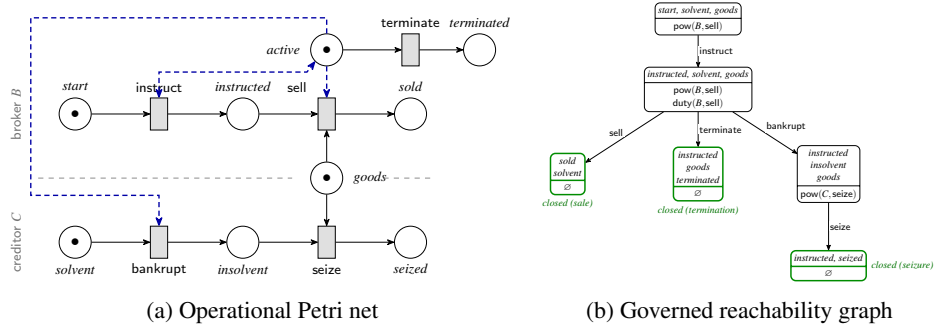
\begin{figure}[t]
\centering

\begin{subfigure}[t]{0.55\textwidth}
 \centering
 \resizebox{\linewidth}{!}{%
   \begin{tikzpicture}[
  place/.style={circle,draw,minimum size=6.5mm,inner sep=1pt},
  trans/.style={rectangle,draw,fill=black!12,minimum width=3.5mm,minimum height=6.5mm},
  plabel/.style={font=\scriptsize\itshape},
  tlabel/.style={font=\scriptsize\sffamily},
  amend/.style={blue!65!black,thick,densely dashed},
  >={Stealth[round]}
]
  \node[place] (start) at (0,1.3) {$\bullet$};
  \node[trans] (instruct) at (1.7,1.3) {};
  \node[place] (instructed) at (3.4,1.3) {};
  \node[trans] (sell) at (5.1,1.3) {};
  \node[place] (sold) at (6.8,1.3) {};
  \node[place] (goods) at (5.1,0) {$\bullet$};

  \node[place] (solvent) at (0,-1.3) {$\bullet$};
  \node[trans] (bankrupt) at (1.7,-1.3) {};
  \node[place] (insolvent) at (3.4,-1.3) {};
  \node[trans] (seize) at (5.1,-1.3) {};
  \node[place] (seized) at (6.8,-1.3) {};

  \node[plabel] at (0,1.85) {start};
  \node[tlabel] at (1.7,1.85) {instruct};
  \node[plabel] at (3.4,1.85) {instructed};
  \node[tlabel] at (4.5,1.85) {sell};
  \node[plabel] at (6.8,1.85) {sold};
  \node[plabel,anchor=west,xshift=2.5pt] at (goods.east) {goods};

  \node[plabel] at (0,-1.85) {solvent};
  \node[tlabel] at (1.7,-1.85) {bankrupt};
  \node[plabel] at (3.4,-1.85) {insolvent};
  \node[tlabel] at (5.1,-1.85) {seize};
  \node[plabel] at (6.8,-1.85) {seized};

  \node[place] (active) at (5.1,2.6) {$\bullet$};
  \node[trans] (terminate) at (6.8,2.6) {};
  \node[place] (terminated) at (8.2,2.6) {};
  \node[plabel,anchor=east,xshift=-2.5pt] at (active.west) {active};
  \node[tlabel] at (6.8,3.15) {terminate};
  \node[plabel] at (8.2,3.15) {terminated};

  \draw[->] (start)--(instruct);
  \draw[->] (instruct)--(instructed);
  \draw[->] (instructed)--(sell);
  \draw[->] (sell)--(sold);
  \draw[->] (goods)--(sell);
  \draw[->] (goods)--(seize);
  \draw[->] (solvent)--(bankrupt);
  \draw[->] (bankrupt)--(insolvent);
  \draw[->] (insolvent)--(seize);
  \draw[->] (seize)--(seized);
  \draw[->] (active)--(terminate);
  \draw[->] (terminate)--(terminated);

  \draw[->,amend] (active)--(sell);
  \draw[<->,amend]
    (instruct.north) -- (1.7,2.15) -- (4.7,2.15) -- (active.south west);
  \draw[->,amend]
    (active.north) -- (5.1,3.25) -- (-0.9,3.25) -- (-0.9,-0.35)
    -- (1.7,-0.35) -- (bankrupt.north);

  \draw[dashed,black!45] (-0.75,0)--(4.4,0);
  \draw[dashed,black!45] (6.25,0)--(7.3,0);
  \node[font=\scriptsize\sffamily,black!55,rotate=90]
    at (-1.2,1.3) {broker $B$};
  \node[font=\scriptsize\sffamily,black!55,rotate=90]
    at (-1.2,-1.3) {creditor $C$};
\end{tikzpicture}%
 }
 \caption{Operational Petri net}
 \label{fig:net-fixed}
\end{subfigure}
\hfill
\begin{subfigure}[t]{0.43\textwidth}
 \centering
 \resizebox{\linewidth}{!}{%
   \begin{tikzpicture}[
  cfg/.style={
    rectangle split,
    rectangle split parts=2,
    draw,
    rounded corners,
    font=\footnotesize,
    inner sep=2.5pt,
    align=center
  },
  good/.style={cfg,very thick,draw=green!55!black},
  misal/.style={cfg, very thick, draw=red!75!black},
  elbl/.style={font=\scriptsize\sffamily,fill=white,inner sep=1pt},
  >={Stealth[round]}
]
    \node[
      cfg,
    ] (m0) at (0,4.6)
      {\textit{start, solvent, goods}
       \nodepart{two} $\pw(B,\mathsf{sell})$};

  \node[cfg] (m1) at (0,2.6)
    {\textit{instructed, solvent, goods}
     \nodepart{two} $\pw(B,\mathsf{sell})$\\ $\duty(B,\mathsf{sell})$};

  \node[
    good,
    label={[font=\scriptsize\itshape,green!45!black]below:closed (sale)}
  ] (m2) at (-3.7,0.2)
    {\textit{sold}\\ \textit{solvent}
     \nodepart{two} $\varnothing$};

  \node[
    good,
    label={[font=\scriptsize\itshape,green!45!black]below:closed (termination)}
  ] (m3) at (0,0.2)
    {\textit{instructed}\\ \textit{goods}\\ \textit{terminated}
     \nodepart{two} $\varnothing$};

  \node[cfg] (m4) at (3.7,0.2)
    {\textit{instructed}\\ \textit{insolvent}\\ \textit{goods}
     \nodepart{two} $\pw(C,\mathsf{seize})$};

  \node[
    good,
    label={[font=\scriptsize\itshape,green!45!black]right:closed (seizure)}
  ] (m5) at (3.7,-2.4)
    {\textit{instructed, seized}
     \nodepart{two} $\varnothing$};

  \draw[->] (m0) -- node[elbl,right] {instruct} (m1);
  \draw[->] (m1) -- node[elbl,above left,pos=.5] {sell} (m2);
  \draw[->] (m1) -- node[elbl,right] {terminate} (m3);
  \draw[->] (m1) -- node[elbl,above right,pos=.5] {bankrupt} (m4);
  \draw[->] (m4) -- node[elbl,right] {seize} (m5);
\end{tikzpicture}%
 }
 \caption{Governed reachability graph}
 \label{fig:reach-fixed}
\end{subfigure}

\caption{Repaired net and partial governed reachability graph.
Amended arcs are dashed and blue. The transitions \textsf{sell},
\textsf{terminate}, and \textsf{bankrupt} consume \textit{active};
\textsf{instruct} reads it. With the amended policy, every reachable
configuration is aligned. Only the initial \textsf{instruct} branch
and selected state components are shown.}

\label{fig:corrected-net-and-reachability}
\end{figure}

While alignment by itself can help identify degenerate normative designs, it should not be considered as the only repair target (\eg the \emph{empty} net is vacuously aligned). 
A repair must preserve a designer-given set $\mathit{Goal}$ of outcome
conditions. 
For every $g\in\mathit{Goal}$, there must be a terminal
configuration $(M,I)\in\RC(\mathcal{G}')$ such that $g(M,I)$ holds.
In the example, the three conditions require $M(\mathit{sold})>0$,
$M(\mathit{terminated})>0$, or $M(\mathit{seized})>0$, respectively.
Synthesising a \emph{maximal} aligned restriction subject to this constraint is related to supervisory control \cite{ramadge1987supervisory}; we leave the formulation and complexity of this synthesis problem to future work and exhibit a repair by hand (visualised in Fig.~\ref{fig:corrected-net-and-reachability}). 
We apply a sequence of institutional and operational repairs, keeping $M_0$ fixed.
The \emph{ungrounded power} $\pw(B,\mathsf{sell})$ outliving the \textit{goods} is repaired \emph{institutionally} by adding the reactive rule $(\mathsf{sell},\mathsf{deact},\pw(B,\mathsf{sell}))$. 
In this way, the power lifecycle is pinned to the consumable goods rather than left to a separate institutional act. 
The \emph{illegitimate ability} of $B$ retaining \textsf{sell} after \textsf{bankrupt} has revoked the authority is repaired operationally by removing the ability: we make \textsf{bankrupt} consume the mandate token \textit{active} and let \textsf{instruct} read \textit{active} so the mandate cannot be re-opened once resolved.
Each of \textsf{sell}, \textsf{terminate}, and \textsf{bankrupt} consumes \textit{active}. 
Accordingly, $\Pi'$ extends $\Pi$ so that each deactivates all initially conferred powers and $\duty(B,\mathsf{sell})$, while \textsf{bankrupt} still activates $\pw(C,\mathsf{seize})$.

\section{Discussion and Future Work}
\label{sec:discussion}

The main goal of our proposed framework is to support the analysis and specification of normative designs for traceable and controllable computational systems.
A process-centred modelling approach to the representation of the operational layer in normative scenarios, in the style of \cite{sileno2017petrinet}, has been only partially explored in previous work.
Governatori and Rotolo~\cite{governatori2008algorithm} present a formal approach to deontic policy specification on Business Process Modelling Notation (BPMN), but do not model its relation to potestative power \ie power as a stateful object bound to operational transitions. 
Ramezani et al.~\cite{ramezani2012compliance} also leverage process modelling formalism in combination with event log analysis to introduce trace-based compliance checking in a purely deontic sense; as such, their contribution only tackles \emph{ex-post} analysis of historical execution. 
In contrast, our solution specifically targets strategies for evaluating a system before or during its deployment.
InstAL~\cite{cliffe2009instal} explicitly models institutional power and binds it to an event-based transition system.
Similarly, contract-oriented frameworks such as Symboleo~\cite{sharifi2020symboleo} do represent potestative positions and their institutional effects.
However, these frameworks lack a process-centred representation that relates institutional powers to independently modelled operational abilities and characterises their alignment throughout execution over consumable resources.
Mapping the normative lifecycle to these operational primitives allows us to expand on the traditional separation of \cite{artikis2009specifying} by grounding institutional positions onto a process state space.
In this context, it is important to remark that our definition of \emph{operational groundedness} (see Def.~\ref{def:grounded}) only expresses possible future exercise, not guaranteed fulfilment or an agent’s ability to ensure it.
Stochastic Petri-net semantics \cite{bause2002stochastic} could support quantitative analysis of the likelihood or timing of future exercise.
However, relating such measures to agentive reliability as explored in \cite{marino2026power} remains future work.
Finally, none of the aforementioned contributions addresses the characterisation of alignment between operational and institutional layers.
The present work offers a mathematically grounded framing for the desirable and undesirable normative states reachable within a given design, and for identifying and tracing misalignments and reasoning about repairs.
Developing efficient verification procedures and synthesising repairs that preserve the intended process outcomes remain directions for future work.

\section{Conclusion}
\label{sec:conclusion}

In this study, we present a formal framework apt to represent operational and institutional states to aid normative system design.
We reduce operational ability to process modelling primitives in order to leverage their concurrent and traceable execution semantics.
Based on this stateful representation of the operational layer, we discuss its relation to the lifecycle of normative positions at the institutional level and present desirable properties that characterise design-level alignment between the two.
Then, taking inspiration from a reported consignment dispute, we show through a stylised scenario how such a process-oriented representation can support the \emph{ex-ante} detection of undesirable normative configurations emerging from the system design.
The framework distinguishes positions whose exercise has become operationally impossible from enabled actions unsupported by institutional power, making both kinds of misalignment explicit over reachable configurations.
We position our solution against the related work in computational settings, by observing how no previous contribution studies the grounding of institutional state onto resource-constrained process execution.
Finally, we discuss future directions leading to the automated detection and eventual repair of misaligned configurations in normative designs.

\paragraph{Acknowledgements.}

This research was partially funded by the SSC-ICT organisation of the Netherlands, which is part of the Ministry of the Interior and Kingdom Relations, and has been carried out as part of the Open Innovation for Verified Automation (OIVA) project, co-funded by the European Union under Grant Agreement No 101298830. Views and opinions expressed are, however, those of the author(s) only and do not necessarily reflect those of the European Union or the European Commission. Neither the European Union nor the European Commission can be held responsible for them.

\bibliographystyle{vancouver.bst}
\bibliography{references}

\end{document}